# Neurons Are All You Need

**Joseph B. Alzagatiti[†]*, Alain C. Glanzman[†]*, Perry A. Demsko[†]*, Pablo I. Salanova[†]*, and David L. Glanzman[1-3]***

*[1]Department of Integrative Biology and Physiology, UCLA College, University of California, Los Angeles, Los Angeles, CA, USA. [2]Department of Neurobiology, and [3] Brain Research Institute, David Geffen School of Medicine at UCLA, Los Angeles, CA, USA. [†]Work performed at UCLA. *All authors contributed equally to this work. Correspondence can be sent to: D.L. Glanzman (glanzman@ucla.edu).*

## Abstract

*For decades, artificial neural networks have operated under the assumption that durable, learned capability resides exclusively in adaptive synaptic parameters, primarily due to the established neurobiology of synaptic connectivity. This design forces models to remain stateless during deployment, incurring well-known limitations such as catastrophic forgetting and expensive retraining. However, recent neurobiological research by Glanzman and colleagues, and others, shows that memories can be transferred between individual organisms in a synaptically independent manner. Inspired by this research, we sought to demonstrate memory transfer of persistent memory between AI model instances, leveraging a neurocentric version of memory. In this paper, we propose NURESA (Neural Unit with Retentive Self-Adaptive Architecture), a novel neural layer that eschews sole reliance on synaptic weight updating by introducing persistent intra-unit memory as a first-class, computational primitive. By structurally separating computation from persistent memory, NURESA enables autonomous memory accumulation and consolidation into transferable stores of memory registers. Experiments across MLP, GPT-2, and modern frontier models like Qwen2.5 demonstrate qualitatively new capability: zero-shot transfer of learned capability into a naive model instance without retraining or access to the original training data; this transfer is enabled by a sparse set of parameters in NURESA's memory register that the model selects on its own based on which neurons were most active. The memory transfer capability of NURESA also protects against domain loss when transformer models undergo learning via fine-tuning. Finally, the memory transfer process completes in approximately one minute even for transformer models with millions to billions of parameters, producing a transfer file orders of magnitude smaller than the original model.*

## 1 Introduction

Conventional neural sequence transduction models and deep learning architectures have firmly established state-of-the-art performance by optimizing adaptive parameters via global backpropagation (Vaswani et al., 2017 [1]; Sutskever et al., 2014 [2]). However, this foundational paradigm imposes a fundamental constraint: learned capability is diffusely locked within weight matrices. As a result, current models remain stateless during deployment; whether a model has interacted with an environment for a year or a second, its underlying weight configuration remains unchanged unless subjected to retraining. Existing attempts to endow models with "memory" typically rely on context window augmentation or global fine-tuning. Methods of context window manipulation in a "memory-like" manner include: prepending user preference stores into the system prompt (e.g., OpenAI's ChatGPT Saved Memories; OpenAI, 2024 [3]), external profile recall (e.g., Google Gemini's Saved Info; Google, 2024 [4]), and/or markdown-based project memory files loaded into prompt prefixes via context caching (e.g., Anthropic Claude's Project Memory; Anthropic, 2024 [5]). Altogether, these approaches do not alter the nature of the computational substrate itself.

Rather than proposing another optimization trick within the parameter-centric paradigm, we ask a more fundamental question: ***What if persistent, neuron-centric memory were a first-class computational primitive?***

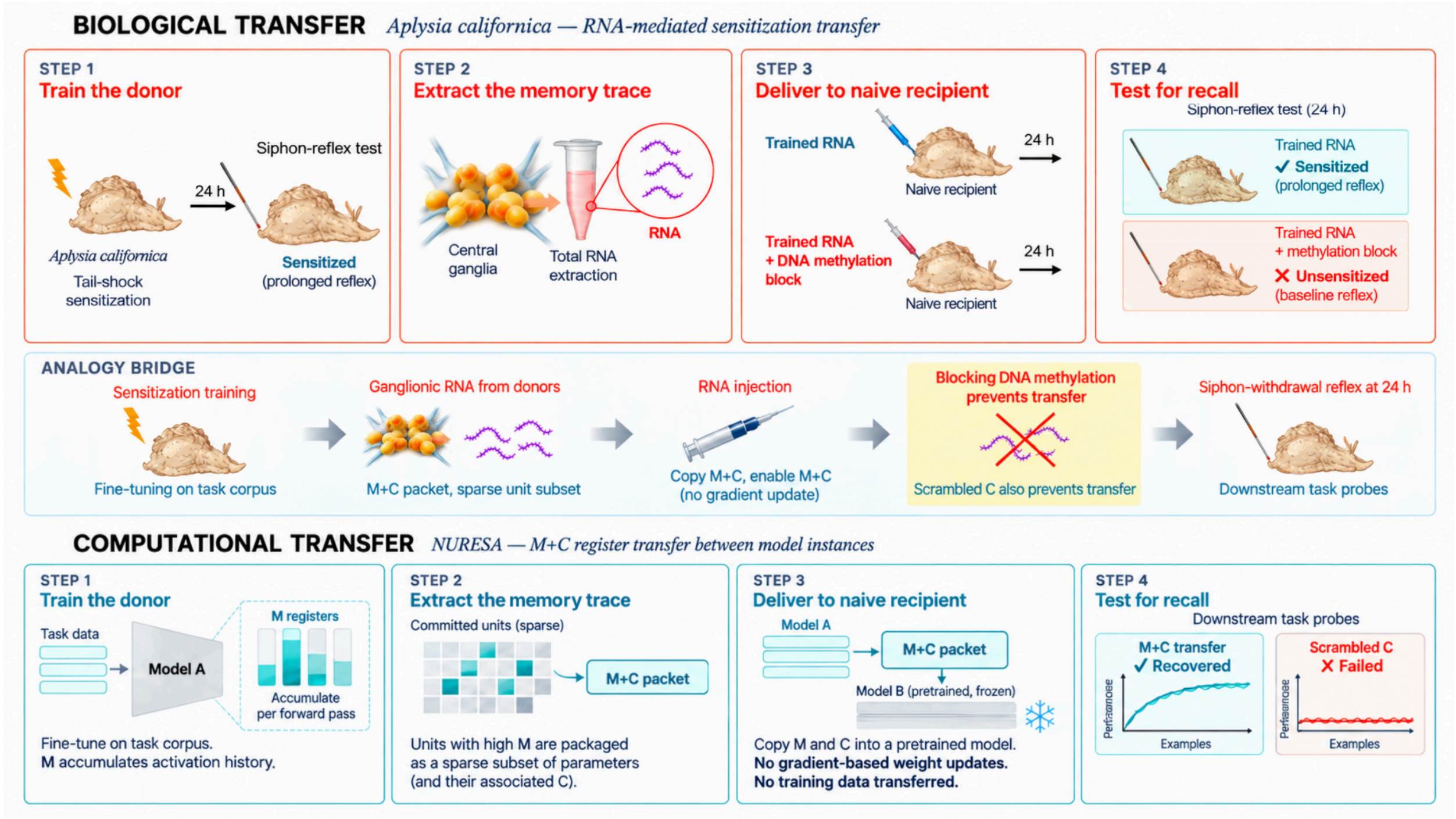

**Figure 1. Memory Transfer Across Substrates.** A schematic on how biological memory transfer can serve as a template in machine learning models for both neuronal memory registers and the subsequent transfer of persistent memory stores.

Here, we introduce NURESA, or Neural Unit with Retentive Self-Adaptive Architecture, an additive, persistent memory layer inspired by the neurobiology of memory transfer that integrates directly into existing neural architectures without replacing backpropagation or transformer layers (**Figure 1**). NURESA assigns each computational unit an append-only internal memory register (M) that tracks activation history, modulates non-linear activation responses, and gates local Hebbian plasticity. When units cross an activation-driven commitment threshold, their learned configurations are consolidated into an independent persistent vault (C).

By decoupling persistent memory from synaptic connectivity, NURESA transforms learned capability from a diffuse property of weight tensors into a discrete, portable, and inspectable object. Experimentally, we showed that the NURESA M+C packet yields memory transfer-like improved capabilities in each model tested (MLP, GPT-2, Qwen2.5) without the need of any post-training of the receiving model. In the case of the modern Qwen2.5 1.5B, we recovered 88.15% of the supervised fine-tuning (SFT) performance in a single domain of GSM8K relatively instantaneously, ranging from 65.1 to 78.0 seconds, up to a 48.9X speed up compared to fresh fine-tuning of the instance with the same settings. Also, we noticed that the M+C packet at the most extreme for Qwen2.5 was 32.7X smaller than the base model, while still retaining SFT improvement. Of significant surprise were two findings: 1) M+C packets mitigated catastrophic forgetting in MATH500 from SFT of Qwen2.5; and 2) M-selected neurons committed into M+C outperformed any other selection of neurons to recover performance gains during M+C memory transfer process, up to 91.5% of the SFT benchmark in GSM8K. Finally, M+C grants privacy for training data from a reconstruction attack for training representations, resembling random selection, as well as membership inference where M+C was at near chance for AUROC at 0.4769. We note similar results in MLP trained for the MNIST task and GPT-2 medium.

Our key contributions are:

- **Model Independent, Stateful Memory Substrate:** NURESA is a computational architecture that incorporates a persistent, intra-unit memory; the memory accumulates and consolidates knowledge as a model learns. This architecture is applicable to radically different artificial intelligence models, for example, MLPs and transformers.
- **Empirical Validation of Model Memory Transfer:** NURESA is capable of zero-shot transfer of consolidated memories between same-architecture models at both unit and model levels, thereby dispensing with gradient updates. In larger transformer models such as Qwen2.5 1.5B, NURESA protects against catastrophic forgetting from fine-tuning.
- **Memory Register Confers Privacy:** The memory register in NURESA prevents a rival agent from reconstructing the content of the training sets used for fine-tuning.

***NOTE: This paper presents the conceptual framework and empirical results of the NURESA architecture. Implementation details, including full protocols, source code, and trained model weights, are subject at present to the request process outlined in the Data and Code Availability section. These details will also be released upon the paper's publication in a peer-reviewed journal.***

# 2 Background and Biological Motivation

## 2.1 Biological Foundations of Non-Synaptic Memory

Mainstream artificial intelligence and neurobiology have shared the long-standing, foundational assumption that long-term memory is encoded exclusively in the connectivity and strength of synaptic junctions (Abraham et al., 2019 [6]; Hassabis et al., 2017 [7]; Mondal et al., 2026 [8]; Zeleznikow-Johnston et al., 2025 [9]). However, recent theoretical considerations (Akhlaghpour, 2022 [10]; Gallistel, 2021 [11]; Gershman, 2023 [12]) and empirical evidence from the field of neuroscience have raised significant questions for the synaptic plasticity and memory (SPM) hypothesis. Among the problematic findings for the SPM hypothesis, investigators have shown that long-term memory (LTM) can survive the significant disruption of learning-induced growth of new synapses in mice that have undergone artificial hibernation or been subjected to retrograde amnesia as well as invertebrate animals in which memory-related synapses have been experimentally erased (Chen et al., 2014 [13]; Lin et al., 2026 [14]; Ryan et al., 2015 [15]). Particularly challenging for the SPM hypothesis, and the foundation of NURESA's memory transfer capability, is the demonstration that ***memory can be transferred between individuals via genetic substrates*** (Bédécarrats et al., 2018 [16]; Moore et al., 2021 [17]).

Also inconsistent with the SPM hypothesis is the discovery that ***memory can be transferred transgenerationally***. Transgenerational inheritance of memory has now been documented in mice and other organisms (Dias & Ressler, 2014 [18]; Zidan et al., 2025 [19]). Interestingly, some of the molecular agents implicated in the transmission of memory across generations are species of non-coding RNA (Posner et al., 2019 [20]; Zidan et al., 2025 [19]), echoing the finding that RNA can mediate memory transfer between animals (Bédécarrats et al., 2018 [16]).

Evidence of learning and memory in single-celled organisms represents a further difficulty for the SPM hypothesis. Most demonstrations of learning in such organisms have involved simple, non-associative forms of memory, specifically habituation and sensitization (Boisseau et al., 2016 [21]; Eisenstein et al., 1982 [22]; Laeverenz-Schlogelhofer & Wan, 2026 [23]; Pla-Mauri & Solé, 2026 [24]; Rajan et al., 2023 [25]; Wood, 1969 [26]). But there are also reports of associative learning, particularly classical conditioning, in protists (unicellular eukaryotes) (Alipour et al., 2017 [27]; Armus et al., 2006 [28]; De la Fuente et al., 2019 [29]; Doan et al., 2026 [30]; Gelber, 1952 [31]; Hennessey et al., 1979 [32]), although some of the reports have been met with skepticism (Gershman et al., 2021 [33]).

There is ample evidence for LTM in single cells. The plasmodial slime mold *Physarum polycephalum*, a non-neural organism, can habituate to an aversive chemical stimulus (quinine or caffeine) and the habituation can persist for up to 2 days (Boisseau et al., 2016 [21]). In addition, if slime molds are given long-term habituation to an aversive chemical (in this case, salt) and then made to enter a dormant form through being desiccated (known as sclerotia), they retain long-term habituation memory even after a month of dormancy (Boussard et al., 2019 [34]). Furthermore, a single slime mold cell can fuse to another slime mold cell when they are placed together. When a habituated cell is fused to a non-habituated cell, the habituation memory is transferred to the untrained cell (Vogel & Dussutour, 2016 [35]). Although the mechanism of memory transfer in slime molds is not known, Vogel and Dussutour's (2016) [35] results suggest the trained cell is transferring what it has learned to the untrained cell through conveyance of one or more molecules. These results in the slime mold echo recent reports of molecular transfer of LTM in the invertebrates *Aplysia californica* (Bédécarrats et al., 2018 [16]) and Caenorhabditis elegans (Moore et al., 2021 [17]). Long-term habituation memory — memory that persists for several hours — has also been shown in the contractile protozoan *Stentor* (Rajan et al., 2026 [36]; Rajan & Marshall, 2025 [37]; Wood, 1969 [26]; see also Ramdas et al., 2026 [38]). Finally, associative LTM has been reported for a unicellular organism. Hennessey et al. (1979) [32] presented evidence that the ciliate *Paramecium* can be classically conditioned by pairing speaker-generated vibrations with electrical shocks. The memory for this form of classical conditioning persisted for 24 hr according to these investigators.

Given that protozoans can exhibit LTM, are the mechanisms of LTM the same or similar in unicellular and multicellular animals (metazoans)? One might imagine that the evolution of neurons approximately 600 million years ago (Sachkova et al., 2025 [39]) ushered in new cellular machinery for storing experience based on synaptic change. If so, then knowledge of how memory is stored in unicellular organisms would be unlikely to yield major insights into the biology of LTM in animals with brains, including humans. But many of the signaling pathways known to play critical roles in LTM in metazoans are present in unicellular organisms. For example, the cyclic adenosine monophosphate (cAMP)-to-protein kinase A (PKA) intracellular cascade, prominently implicated in LTM in both vertebrates and invertebrates (Kandel, 2012 [40]), is present in ciliates (Hochstrasser & Nelson, 1989 [41]) and in *Dictyostelium*, a cellular slime mold (Mann et al., 1997 [42]), among other protozoans.

The evidence reviewed above indicates that unicellular organisms exhibit many of the same forms of memory that animals do. This should not be surprising: the capacity for learning confers enormous selective advantages on an organism and it seems improbable to us that life forms existed on earth for over 3 billion years prior to the emergence of this capacity. Moreover, most of the molecular pathways implicated in memory in metazoans were already present in non-metazoans long before the evolutionary appearance of neurons. Does it seem likely that metazoans simply jettisoned intracellular mechanisms that appear to have subserved memory effectively for billions of years? We think not. We believe, rather, that these primordial non-synaptic memory mechanisms continue to play critical mnemonic roles in brains, including the human brain.

What intracellular mechanisms might mediate memory? Besides, or in addition to, those discussed above, Gold and Glanzman (2021) [43] have proposed that LTM is maintained by nuclear changes within neurons, particularly epigenetic changes and genomic changes (Ginsburg & Jablonka, 2009 [44]; Glanzman, 2024 [45]). These types of changes are evolutionarily ancient and in principle could mediate memory in non-metazoans. For example, epigenetic mechanisms of memory have been demonstrated in yeast (Ge & Brickner, 2024 [46]).

Thus, emerging evidence suggests that the SPM hypothesis is incorrect; in particular, it cannot explain the existence of biological memory in a portable, molecular form nor the persistence of memory following synaptic disruption. We believe that biological LTM is encoded by cellular mechanisms, including epigenetic and genomic mechanisms. NURESA is a computational analog of this insight; it embeds persistent memory directly inside artificial units rather than basing memory on inter-unit weight matrices. Consequently, NURESA enables the transfer of neuronal history between individual models just as biological memories can be moved between individual organisms (**Figure 1**).

## 2.2 Working Memory & External Memory Banks in Machine Learning

In machine learning, augmenting neural architectures with memory has traditionally taken two forms: recurrent hidden states (such as Long Short-Term Memory integrators; Hochreiter & Schmidhuber, 1997 [47]) or coupling a neural controller network to an external addressable memory bank, as in the seminal work on Neural Turing Machines (NTM) (Graves, Wayne, & Danihelka, 2014 [48]).

Drawing on psychological models of human working memory (e.g., Baddeley's central executive; Baddeley, Eysenck, & Anderson, 2009 [49]) and prefrontal cortex dynamics (Goldman-Rakic, 1995 [50]; Hazy, Frank, & O'Reilly, 2006 [51]), the NTM couples a feedforward or recurrent neural network controller to an external N x M memory matrix. Interaction with this global memory bank is mediated by continuous, differentiable read and write heads using a combination of context-based addressing and location-based addressing.

While NTMs significantly enhance performance on algorithmic tasks (such as sequence copying, associative recall, and priority sorting) compared to standard LSTMs, they remain structurally constrained by their centralized design: (1) all read and write operations must route through a central neural controller and a fixed set of attentional heads; (2) accessing memory requires generating, shifting, and normalizing soft attentional distributions over all N memory rows at every time step; and (3) the external RAM matrix in an NTM functions as a temporary scratchpad for episodic variable-binding rather than providing autonomous neuronal-level consolidation or data-free capability transfer.

## 2.3 Decentralized Intra-Unit Memory vs. Central External Memory

NURESA departs fundamentally from external memory paradigms like NTM by decentralizing memory storage into every computational unit, i.e. neuron. Instead of routing signals to an external memory bank via central attentional heads, NURESA endows each neuron with an append-only internal memory register (M) that tracks activation history autonomously on every forward pass. When neurons cross an activation-driven commitment threshold, their learned state consolidates into an intra-unit vault (C) maintained separately from synaptic weights. The notable emergent quality is that M+C as a portable memory register can be transferred from a fine-tuned model to a fresh instance of another model, regardless of architecture type.

**Table 1: Architectural comparison between Neural Turing Machines (centralized external RAM) and NURESA (decentralized persistent intra-unit memory).**

| Architectural Property | Neural Turing Machines | NURESA |
|---|---|---|
| Memory Topography | Centralized external RAM matrix (N x M) | Decentralized intracellular state (M register and C vault per unit) |
| Addressing & Access | Differentiable read/write heads with content and location addressing | Autonomous signal accumulation and local activation blending |
| Learning Mechanism | Centralized backpropagation through time across controller and heads | Local memory-gated Hebbian plasticity without global backpropagation |
| Primary Function | Short-term algorithmic working memory and variable binding | Durable, portable capability storage and enables zero-shot memory transfer |

# 3 NURESA Architecture

NURESA operates as a set of cooperating intra-unit mechanisms: an append-only memory register (M), a memory-modulated activation blend, local memory-gated plasticity, autonomous commitment, cellular consolidation into vault (C), and dual-granularity transfer.

## 3.1 Persistent Memory Register (M)

Each unit maintains a persistent scalar register M that accumulates activation history on every forward pass in proportion to signal magnitude. M is append-only and never reset during training, serving as an unsupervised running record of unit participation. As M grows, it acts as a governor that shifts unit execution from instantaneous input dependency to internal state dependency.

## 3.2 Memory-Modulated Activation

Each unit computes its output by dynamically blending a saturating response and a rectifying response governed by M via a sigmoid gating function. At low M, the unit exhibits rectifying behavior; as memory accumulates, it transitions toward a bounded, saturating regime. This enables individual units to capture temporal dynamics without explicit recurrent feedback loops.

## 3.3 Local Memory-Gated Plasticity

Rather than relying exclusively on global backpropagation, NURESA units undergo local, Hebbian-like weight updates gated by internal memory state. Plasticity gating permits adaptation only when the receiving unit is ready, enabling decentralized learning while safeguarding model stability.

### 3.4 Autonomous Commitment, Cellular Consolidation and Engram Sparsity

When a unit's accumulated M crosses a commit threshold (tuned to select a sparse subset of neuronal units):

- **Vault Population:** Its learned configuration consolidates into a persistent vault C.
- **Iterative EMA Consolidation:** Weights are consolidated iteratively via exponential moving averages across post-commitment steps until convergence.
- **Synaptic Protection and Gradient Scaling:** Synaptic weights are protected from further disruptive updates via a plasticity floor, similar to soft weight freezing in PackNet (Mallya & Lazebnik, 2018 [52]).

### 3.5 Memory Transfer Mechanisms

NURESA supports moving learned capability between same-architecture models at two distinct levels:

- **Neuron/Unit-Level Transfer:** Extracted vault states are injected directly into recipient neurons. The recipient recovers source capability zero-shot, without gradient steps or exposure to source training data.
- **Model/Agent-Level Latent Transfer:** The collective memory of multi-unit populations is encoded into a compact, latent vector packet of M+C, and injected into a recipient model/agent.

## 4 Training and Benchmarking Regimes

Evaluations span three model tiers: Multi-Layer Perceptrons (MLPs) on MNIST vision datasets, GPT-2 language models, and a modern, multi-domain frontier-class architecture, Qwen2.5 1.5B. To isolate NURESA's unique memory primitives from standard parameter optimization, benchmarks compare M+C-transfer against baselines and SFT for each model within a harness run with the same settings and hardware. Additionally, for both GPT-2 and Qwen2.5, we ran privacy experiments to determine if the M+C packets could prevent an adversary from ascertaining what was in the training sets during fine-tuning for both reconstruction attacks and data set membership inference.

### 4.1 MLP with MNIST Training and Benchmarking with NURESA M+C Transfer

- **Model and Data.** Experiments use MNIST (LeCun et al., 1998 [53]) with all ten digit classes: the standard 60,000-image training set and 10,000-image test set. Images are normalized (mean 0.1307, standard deviation 0.3081) and flattened to 784 inputs. The model is a three-layer MLP with 269,322 parameters. Model A is trained with Adam (Kingma & Ba, 2015 [54]), using cross-entropy loss on a single GPU. It reaches 98.0% test accuracy. All conditions are scored by top-1 accuracy on the full test set, where chance is 10%.
- **NURESA M+C Memory Transfer Procedure.** The M+C transfer packet is written into Model B, a newly built model with no training. We do an increasing sweep for M+C of the top performing neurons committed to M+C, ranging from 10 to 50%. We also compute the time it takes to do the NURESA overlay for each condition and the file size of each M+C packet. We compare several conditions, including: Model A, the trained reference; Model B before injection, the untrained baseline; Model B with intact M+C; Model B with M+Random C, where C is randomized by being replaced with standard normal noise; and Model B with M+Shuffled C, where C is shuffled by being replaced with permutations of committed neurons.

## 4.2 GPT-2 Training and Benchmarking with NURESA M+C Transfer

- **Model and Data.** All experiments use GPT-2 Medium (345M parameters; Radford et al., 2019 [55]), initialized from the public pretrained checkpoint on HuggingFace (Radford et al., 2019 [55]). The source model (Model A) is fine-tuned on WikiText-103 (Merity et al., 2017 [56]). We stream the raw training split, keep the first 1,800 passages longer than 50 characters, and split them into 1,600 training and 200 held-out evaluation passages. Sequences are truncated or padded with EOS to 128 tokens. Model A is trained for 500 steps. We evaluate with three zero-shot measures: 1) WikiText Perplexity on the 200 held-out WikiText-103 passages. 2) LAMBADA (Paperno et al., 2016 [57]; OpenAI variant, first 500 test examples), where accuracy is top-1 greedy prediction of the first subword token of the passage's final word, given the preceding context. 3) HellaSwag (Zellers et al., 2019 [58]; first 400 validation examples), where each of the four candidate endings is scored by its length-normalized log-likelihood under the context, and the highest-scoring ending is chosen.
- **NURESA M+C Transfer Procedure.** Each M+C wraps transformer submodules, maintaining per-neuron memory and commitment states. We do an increasing sweep for M+C of neurons committed to M+C, ranging from 10 to 50%. We also compute the time it takes to do the NURESA overlay for each condition and the file size of each M+C packet. For transfer, we copy M+C from Model A into a freshly loaded, pretrained GPT-2 Medium (Model B). No gradient updates are applied to Model B. We compare several conditions, including: 1) Model A trained; 2) Model B and C transfer; 3) Model B and random C: Gaussian noise scaled to the per-layer standard deviation of A's C, with the true M; 4) Model B and shuffled C: correct registers assigned to the wrong layers via a cyclic shift of 12 layers; 5) and Vanilla Model B, the lower bound.
- **Privacy Testing of NURESA M+C:** To test whether the transferred package exposes training representations, we ran a gradient-based reconstruction attack on GPT-2 medium after SFT as above, but selected only 30% commitment in the M+C overlay for this experiment. The attack targeted the committed rows of the layer-12 feed-forward block. Following the activation-maximization approach used in feature visualization and model inversion (Erhan et al., 2009 [59]; Fredrikson et al., 2015 [60]), we started from a random Gaussian vector. We then used Adam to maximize the mean absolute projection of x onto the committed rows. We computed the cosine similarity between the optimized x and each of 400 real training representations. These are layer-12 hidden states, mean-pooled over non-padding tokens. We recorded the highest of these similarities, i.e., the closest match to any training example. This was repeated for 50 random initializations. Two conditions were run, the attack on a GPT-2 instance with M+C active, and the attack on 50 unoptimized random Gaussian vectors scored the same way. If the M+C scores match the random baseline, the package retains little recoverable training structure under this attack. To test whether the transferred package reveals which examples were used in training, we ran a membership inference attack (Shokri et al., 2017 [61]; Salem et al., 2019 [62]). We used 1,600 training sequences (members) and 1,600 held-out sequences from the same corpus (non-members). For each sequence, we took the layer-12 hidden state of the trained model, mean-pooled over non-padding tokens, and projected it onto the committed rows of that layer. From this projection we computed summary features, such as the mean, standard deviation, maximum, and minimum of the projection. We trained a logistic regression classifier (scikit-learn; Pedregosa et al., 2011 [63]) on standardized features from a random 70% of the examples and evaluated it on the remaining 30%. Performance is reported as the area under the ROC curve. AUROC equals the probability that a randomly chosen member receives a higher membership score than a randomly chosen non-member. An AUROC of 0.5 corresponds to chance, and 1.0 to perfect calls. If the M+C-derived features yield an AUROC near 0.5 while the full-model features yield a clearly higher AUROC, then membership is detectable from the full model but not from the M+C package.

### 4.3 Qwen2.5 Training and Benchmarking with NURESA M+C Transfer

- **Model and Data.** All experiments use the base Qwen2.5-1.5B checkpoint (1.5B parameters, 28 layers; Qwen Team, 2024 [64]), not the Instruct variant. Models are loaded in bfloat16 and fine-tuned with full-parameter updates, without LoRA or QLoRA. Evaluation uses the EleutherAI Language Model Evaluation Harness (Gao et al., 2024 [65]) with batch size 1 and no example limits. GSM8K is the primary measure of math gains; the other benchmarks measure retention of the base model's broader abilities, including MATH-500, ARC-Challenge, HellaSwag, MMLU, TruthfulQA, HumanEval, and TheoremQA (Qwen Team, 2024 [64]). GSM8K benchmarking for all experimental conditions was 0-shot in contrast to the Qwen's team 4-shot benchmarking (Qwen Team, 2024 [64]).
- **NURESA M+C Transfer Procedure.** During math SFT, M+C overlays were built similar to MLP and GPT-2 model procedures. We do an increasing sweep for M+C of neurons committed to M+C, ranging from 10 to 50%. We also compute the time it takes to do the NURESA overlay for each condition and the file size of each M+C packet. We selected math SFT due to, in our testing, it produced a robust positive gain on GSM8K over the base model. We compare three configurations: base, the unchanged pretrained model; math SFT; and M+C math overlays, a fresh base model with the M+C packet applied without further training. For experiments on M-selectivity, we favored only the 30% commitment overlay, and either had standard M-selection, chose 30% of neurons randomly, across 3 seeds, or chose 30% of those neurons with the largest activation signal.
- **Privacy Testing of NURESA M+C:** Similar to the GPT-2 methodology with key differences noted: for the gradient-based reconstruction attack on Qwen2.5 1.5B after SFT as above, we selected only 30% commitment in the M+C overlay for this experiment. For Qwen2.5, the attack targeted the committed rows of the layer-14 feed-forward block. Following the activation-maximization approach used in feature visualization and model inversion (Erhan et al., 2009 [59]; Fredrikson et al., 2015 [60]), we started from a random Gaussian vector. We then used Adam to maximize the mean absolute projection of x onto the committed rows. For Qwen2.5, we computed the cosine similarity between the optimized x and each of 200 real training representations. These are layer-14 hidden states, mean-pooled over non-padding tokens. We recorded the highest of these similarities, i.e., the closest match to any training example. This was repeated for 50 random initializations. Two conditions were run, the attack on a Qwen2.5 instance with M+C active, and the attack on 50 unoptimized random Gaussian vectors scored the same way. If the M+C scores match the random baseline, the package retains little recoverable training structure under this attack. To test whether the transferred package reveals which examples were used in training, we ran a membership inference attack (Shokri et al., 2017 [61]). For Qwen2.5, members and non-members were drawn from the same shuffled GSM8K training split, and formatted identically: the first 800 examples were used for fine-tuning (members), and the next 800 were never trained on (non-members). For each example, we took the layer-14 hidden state of the trained model, mean-pooled over non-padding tokens, and projected it onto the committed rows of that layer. From this projection we computed summary features, such as the mean, standard deviation, maximum, and minimum of the projection. We trained a logistic regression classifier (scikit-learn; Pedregosa et al., 2011 [63]) on standardized features from a random 70% of the examples and evaluated it on the remaining 30%. Performance is reported as the area under the ROC curve. AUROC equals the probability that a randomly chosen member receives a higher membership score than a randomly chosen non-member. An AUROC of 0.5 corresponds to chance, and 1.0 to perfect calls. If the M+C-derived features yield an AUROC near 0.5 while the full-model features yield a clearly higher AUROC, then membership is detectable from the full model but not from the M+C package.

# 5 Results

## 5.1 Zero-Shot Capability Transfer Using M+C in Different Models

***We demonstrated that an increasing NURESA M+C sweep yields corresponding increases in capability transfer to each model tested (MLP, GPT-2, Qwen2.5) without the need for any training or fine-tuning post-transfer.***

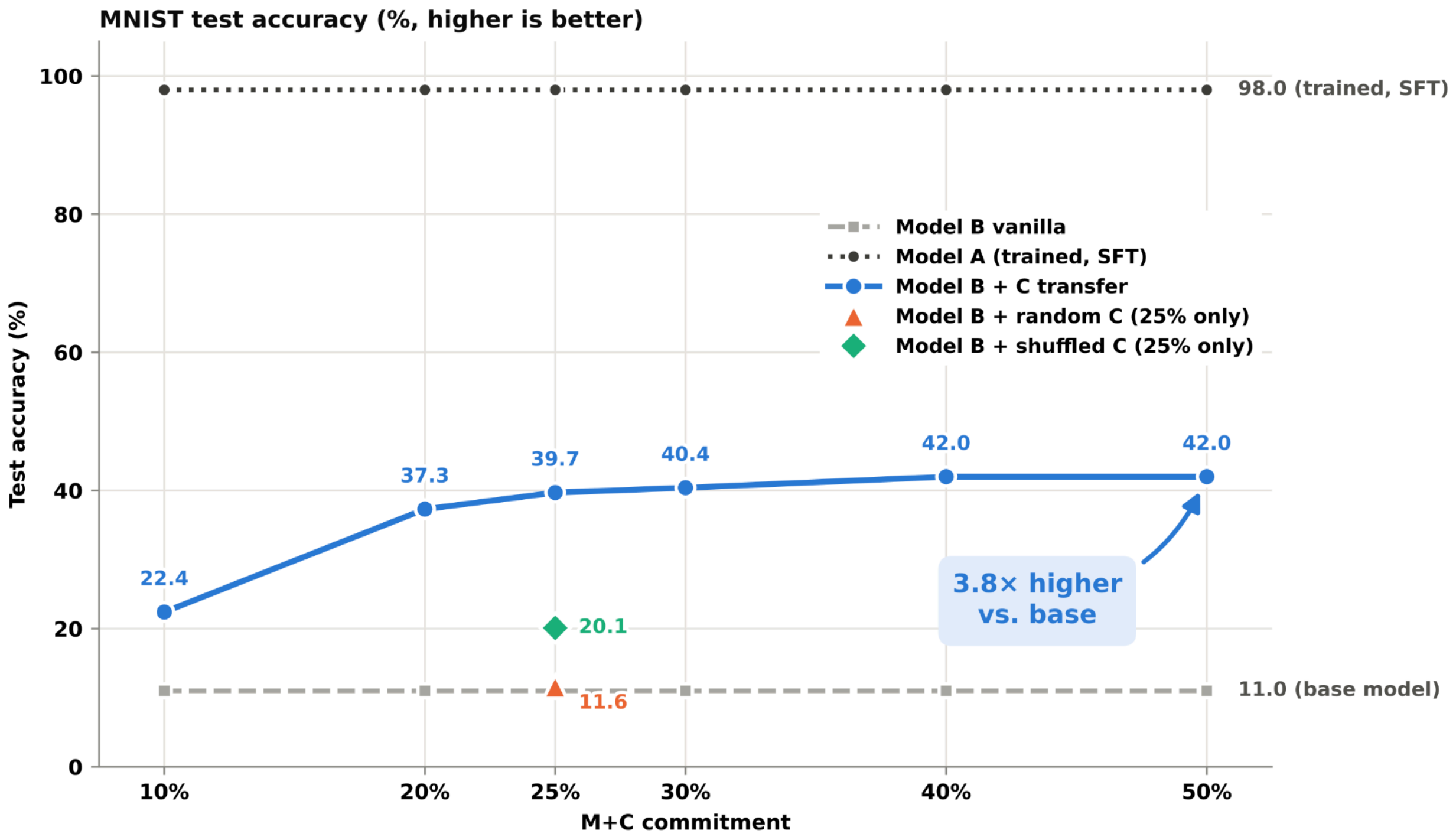

**Figure 2. M+C Zero-Shot Transfers Performance in MLP for MNIST Classification Task.** Increasing M+C yields improved classification performance on a 10 class MNIST task, up to ~3.8X from the untrained base. Different overlay injections are color-coded per the inset legend.

- **M+C Sweep and Zero-Shot Capability Transfer in MLP.** To determine the range of memory transfer fidelity with M+C for a MLP on MNIST data sets, we transferred M+C packets of various neuronal commitments, ranging from 10-50%, into a fresh instanced MLP. We observed that top-1 classification accuracy on the full MNIST test set increasingly scaled with commitments from 22.4% at 10% to 42% at 50% (**Figure 2**). The 50% result is a ~3.8X increase versus Model B's base of 11%, but remains less than the ceiling of a trained, source model's accuracy of 98%. Of note is that ~30% commitment into the M+C transfer overlay represents a key pivot point of capturing almost 2X increase in accuracy, echoing research on core engrams and neural workspaces in large models (Frankle & Carbin, 2019 [66]; Anthropic, 2026 [67]). Finally, at a single test point, random and shuffled C in the M+C overlays caused large degradations in performance, indicating meaningful structure in the transferred state (**Figure 2**).

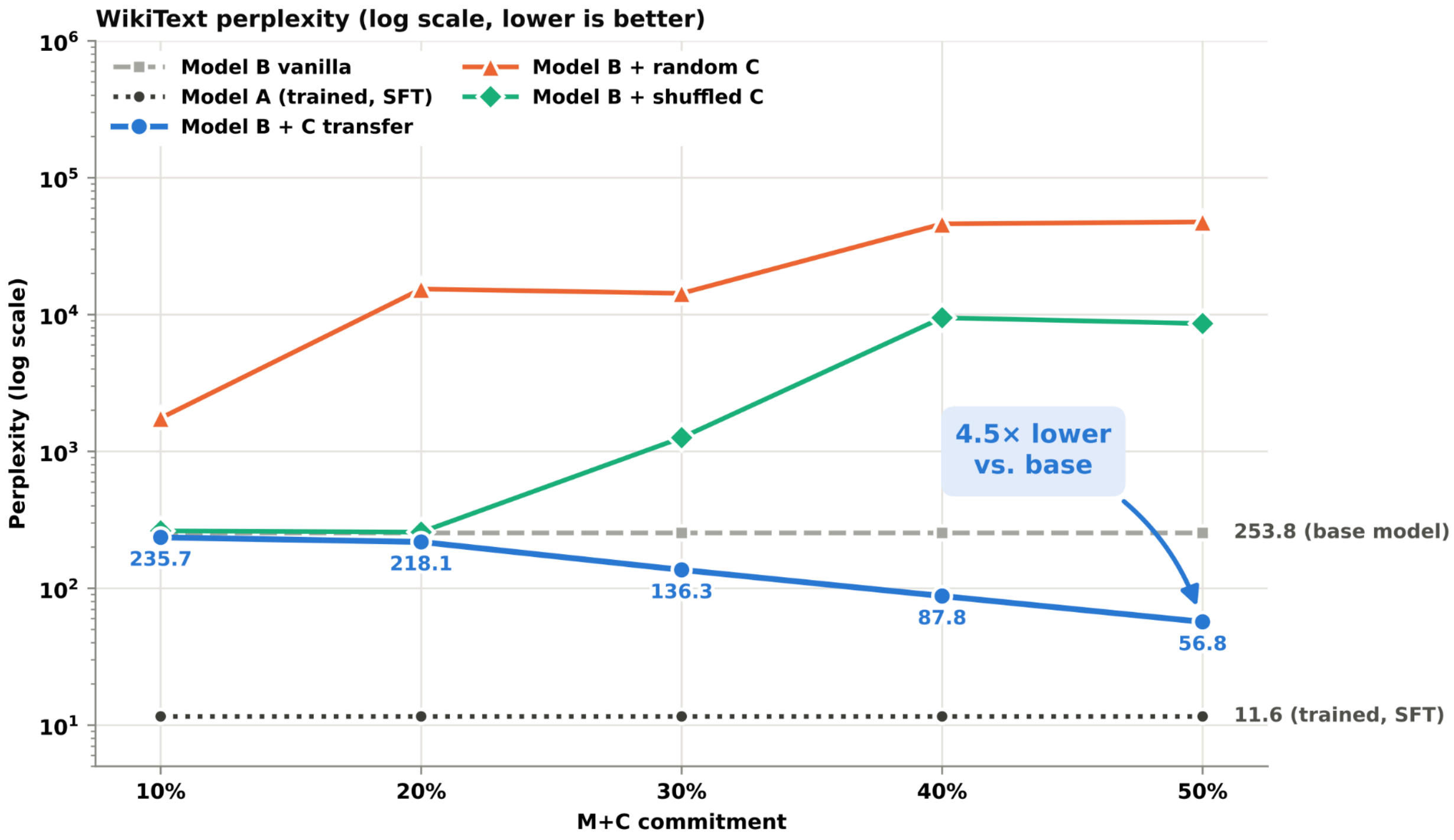

**Figure 3. M+C Zero-Shot Transfers Performance in GPT-2 for WikiText Perplexity.** Increasing M+C yields improved perplexity on the WikiText benchmark for GPT-2, medium, up to ~4.5X from the untrained base. Different overlay injections are color-coded per the inset legend.

- **M+C Sweep and Zero-Shot Capability Transfer in GPT-2.** To determine if M+C can also transfer capability in transformer architectures, we transferred M+C packets of various neuronal commitments, ranging from 10-50%, into a fresh instance of GPT-2 (medium). We observed that WikiText perplexity falls monotonically from 235.71 at 10% to 56.83 at 50% (**Figure 3**). The 50% result is a ~4.5X reduction versus Model B's base of 253.85, but remains less than the ceiling of a trained source model's perplexity of 11.58. As with the MLP demonstration, the 30% commitment M+C transfer overlay represents a key pivot point of capturing almost 2X reduction in the perplexity. For GPT-2, we ran a full sweep of random and shuffled C in the M+C overlays, confirming large degradations in performance, indicating meaningful structure in the transferred state (**Figure 3**).

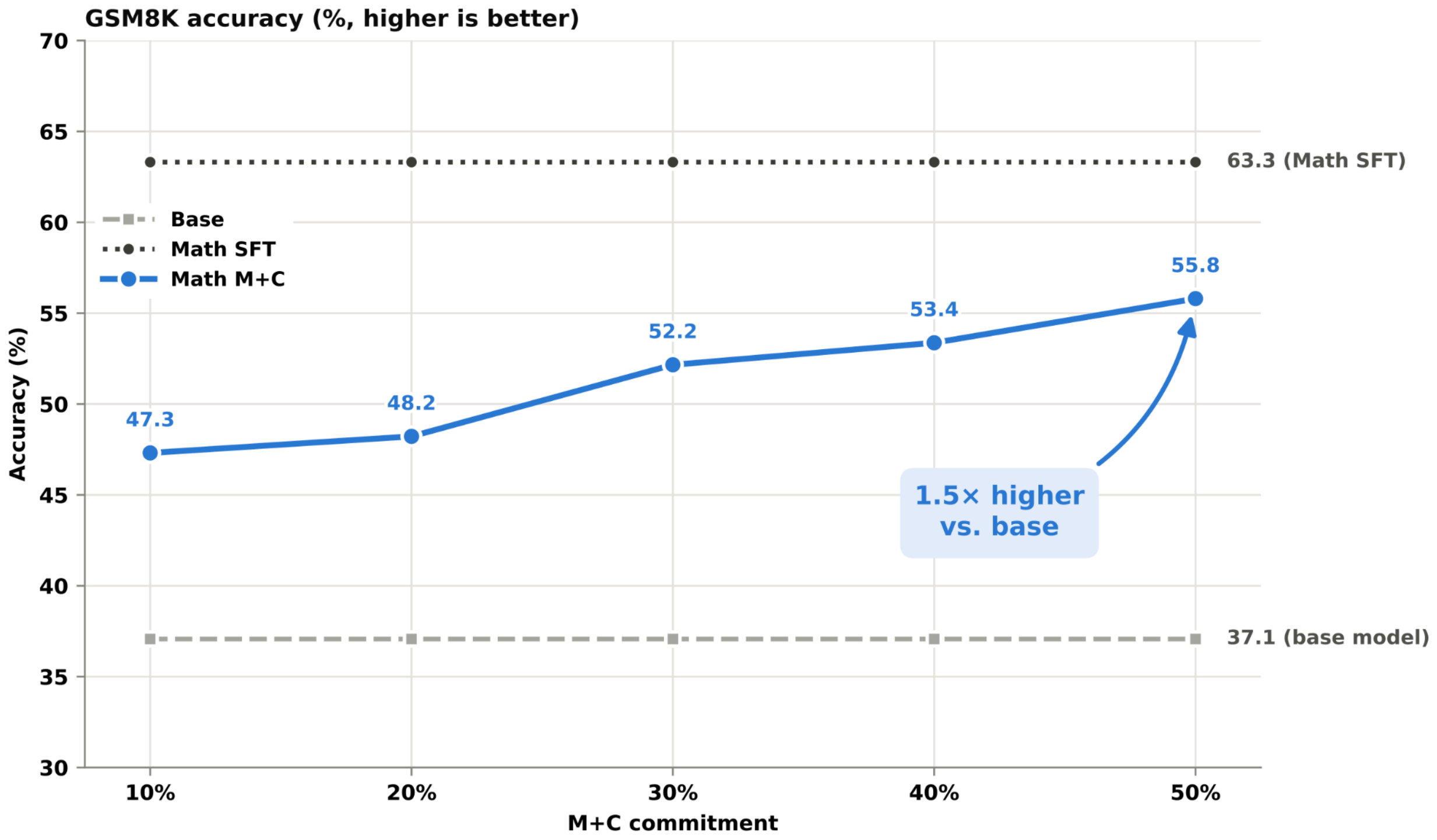

**Figure 4. M+C Zero-Shot Transfers Performance in Qwen2.5 for GSM8K.** Increasing M+C yields improved 0-shot accuracy of Qwen2.5 1.5B on the GSM8K benchmark, up to ~1.5X from the untrained base. Different experimental conditions are color-coded per the inset legend.

- **M+C Sweep and Zero-Shot Capability Transfer in Qwen2.5.** We moved to a multi-domain model in Qwen2.5 1.5B as a balance of a rapidly, tunable model for testing with more modern model features like RoPE, Grouped Query Attention, RMSNorm, and SwiGLU activations, collectively departing in every major mechanism as compared to GPT-2 (Qwen Team, 2024 [64]). Again, to ascertain the fidelity of M+C memory transfer, we transferred M+C packets of various neuronal commitments, ranging from 10-50%, into a fresh instance of Qwen2.5 1.5B. We observed that GSM8K 0-shot benchmark accuracy increases from 47.31 at 10% to 55.8 at 50% (**Figure 4**). The 50% result is up to 1.5X (88.15% of the SFT benchmark) of the untrained base model in the same harness test settings. Of note is that 30% commitment here also represents a key pivot point where the transfer begins to saturate with 40% and 50% commitments having a GSM8K accuracy within ~1-3% points of 30% commitment, reflecting a feature of sub-selection for the NURESA M+C memory transfer register.

## 5.2 M Autonomously Commits High Performing Neurons into M+C Register

Throughout the experimentation with all three models, MLP, GPT-2, and Qwen2.5, we noticed that M+C effectively captured performance as we scaled the number of neurons committed. Recent work by Cammarata and colleagues at Goodfire conjectured that neurons might be less tangled in their organization wherein they organize information according to coherent principles that can be identified (Cammarata, 2026 [68]). We hypothesize that M, which contributes to which neurons to commit to the M+C packet, may be autonomously selecting a subset of the best neurons to retain performance increases.

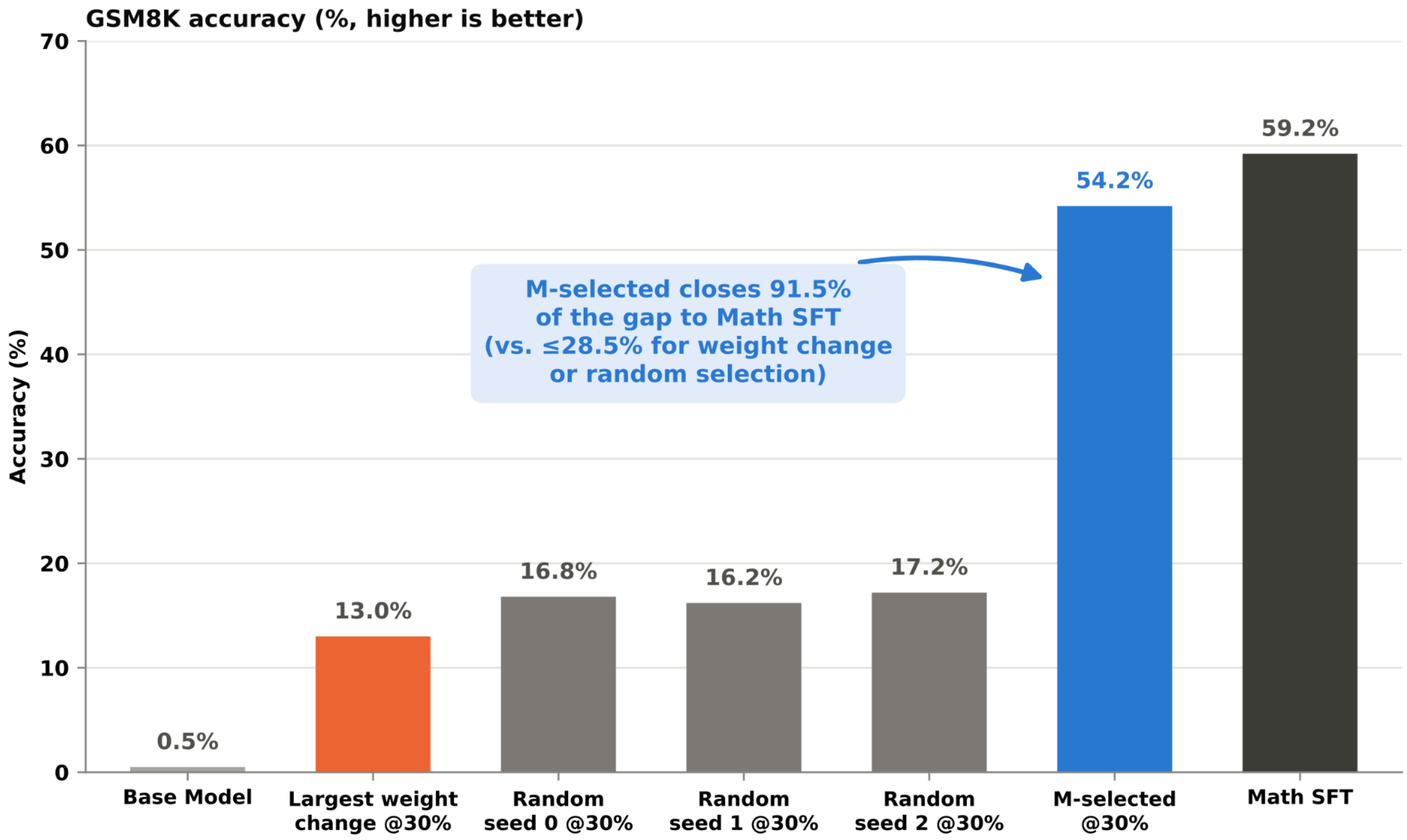

**Figure 5. M-Selected Neurons for M+C Overlays Are High Performing in Qwen2.5 for GSM8K Task.** Autonomously selected neurons by the M tuned threshold recovered 0-shot accuracy of Qwen2.5 1.5B on the GSM8K benchmark up to 91.5% of the SFT performance, while other groups can only recover up to ~28.5%. Different experimental conditions are color-coded per the inset legend.

Using Qwen2.5 as a modern model example with GSM8K performance showing clear M+C overlay success, we asked whether at 30% neuronal commitment M-selected neurons, neurons with largest activation, or three random seeds/selections extracted into a M+C packet could recover performance. In **Figure 5**, we note that ***only the M-selected group for the M+C overlay matched ~91.5% of the SFT trained benchmark on GSM8K***.

## 5.3 M+C Can Endow Domain Protection Against Catastrophic Forgetting

***Across both transformer models we tested, GPT-2 and Qwen2.5, we noted that M+C negated catastrophic forgetting induced by SFT of the models, up to 9 percentage points in the case of Qwen2.5.***

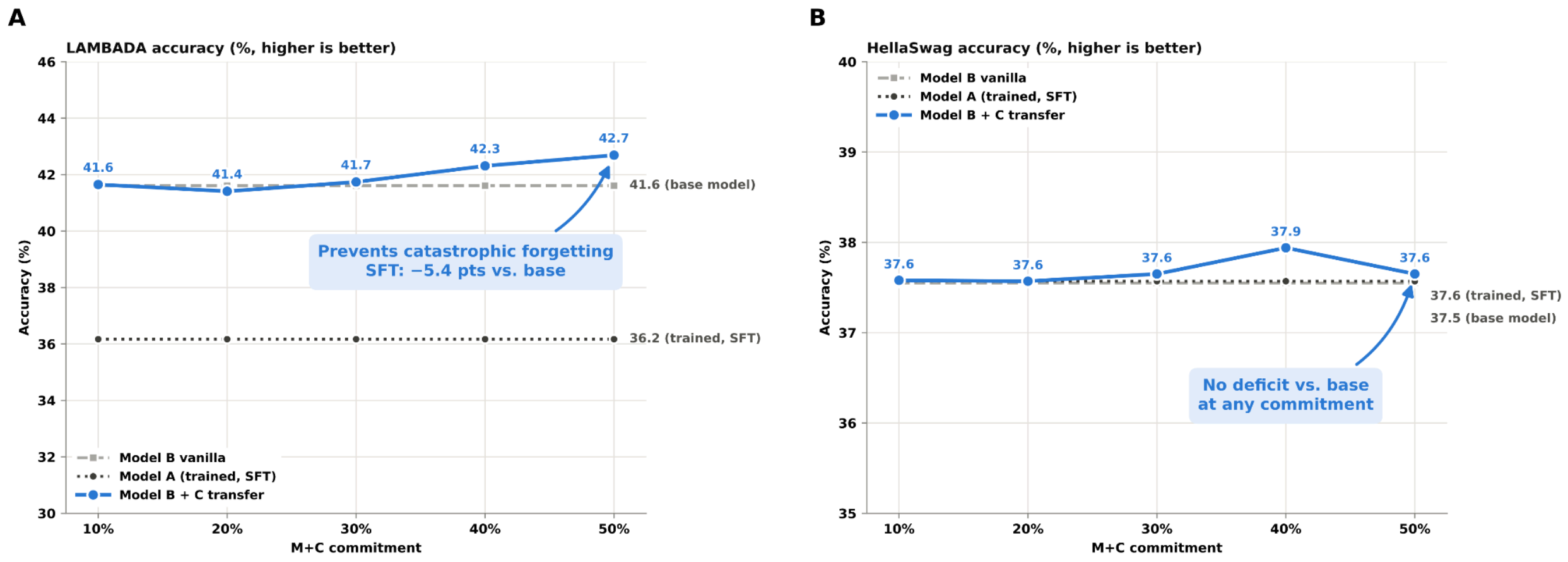

**Figure 6. M+C Protects Against Catastrophic Forgetting from SFT in GPT-2. A)** M+C packets at every commitment level prevent performance loss ~5% from supervised fine-tuning in LAMBADA benchmarks. **B)** M+C overlays did not affect HellaSwag performance. Different overlay injections are color-coded per the inset legend.

- **M+C Protects Select Domains Against Catastrophic Forgetting in GPT-2.** During the M+C sweep, when we analyzed the other benchmarks of GPT-2, medium, we noticed that LAMBADA scores in the NURESA M+C transfers maintained their values as opposed to the trained model experiencing learning-induced forgetting of ~5% points (**Figure 6A**), indicating at this less capable LLM model that NURESA may have catastrophic forgetting protection capabilities. For HellaSwag, M+C did not produce a noticeable effect on the model (**Figure 6B**).

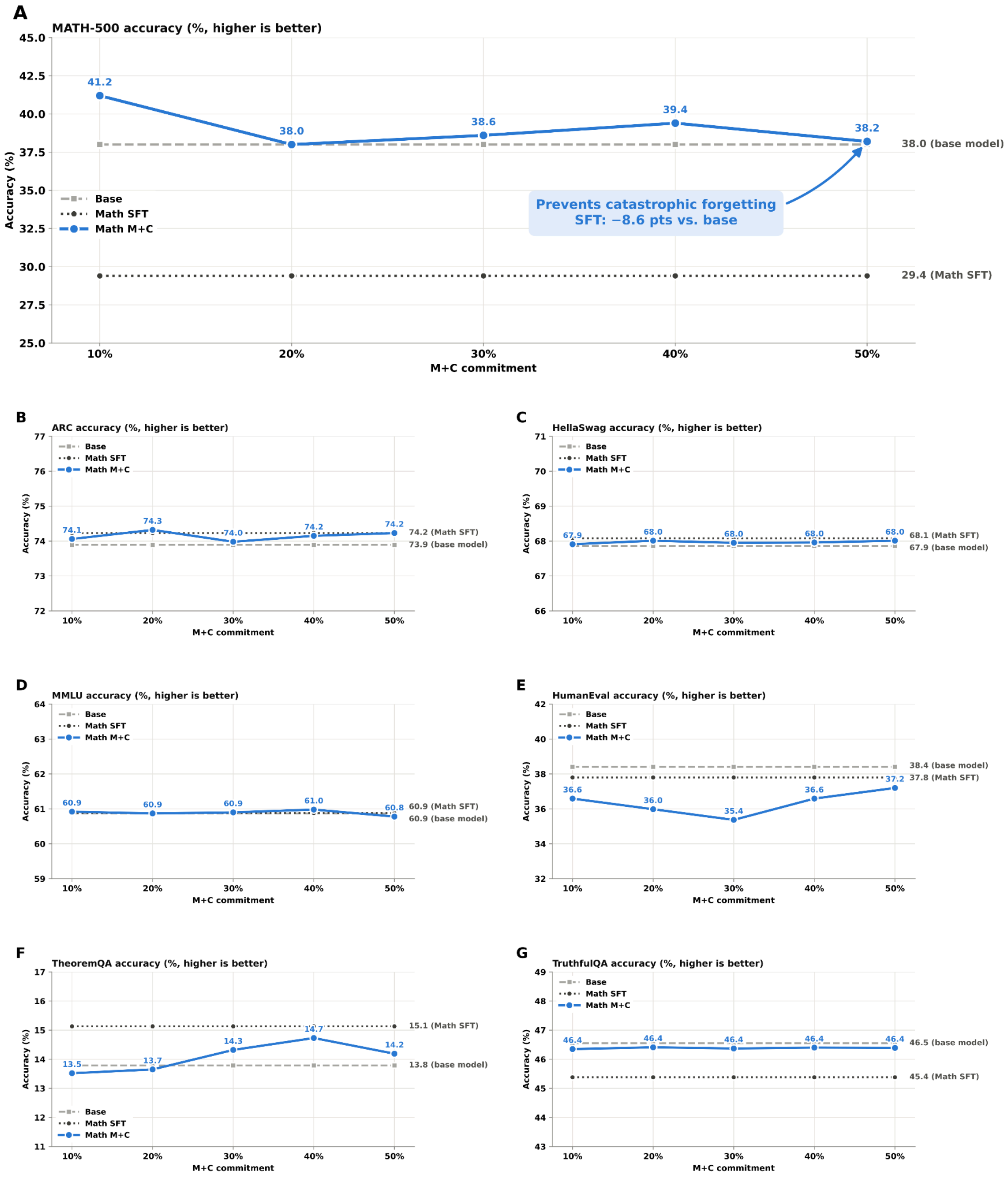

**Figure 7. M+C Protects Against Catastrophic Forgetting from SFT in Qwen2.5. A)** M+C packets at every commitment level prevent performance loss ~9% from supervised fine-tuning in MATH500 benchmarks. **B-G)** M+C overlays did not significantly affect accuracy on *ARC, HellaSwag, MMLU, HumanEval, TheoremQA, or TruthfulQA* benchmark tasks, respectively. Different overlay injections are color-coded per the inset legend.

- **M+C Protects Select Domains Against Catastrophic Forgetting in Qwen2.5.** We also noted that MATH500 scores in the ***NURESA M+C transfers maintained their values as opposed to the trained model experiencing catastrophic forgetting of ~9% points*** (**Figure 7A**), indicating that multi-domain LLM models can be protected by NURESA from catastrophic forgetting. Across all other domains for Qwen2.5, M+C had no significant negative effects on the model's base capabilities (**Figure 7B-G**).

### 5.4 M+C Overlays Compress Model Learning and are Fast to Apply

***Across all three models of MLP, GPT-2, and Qwen2.5, M+C packets are a relatively lightweight and rapid overlay memory application to any model instance - the transformer-based models have, on average, up to ~31X reduction in size to the full models, and up to ~49X faster than a single run of fine-tuning.***

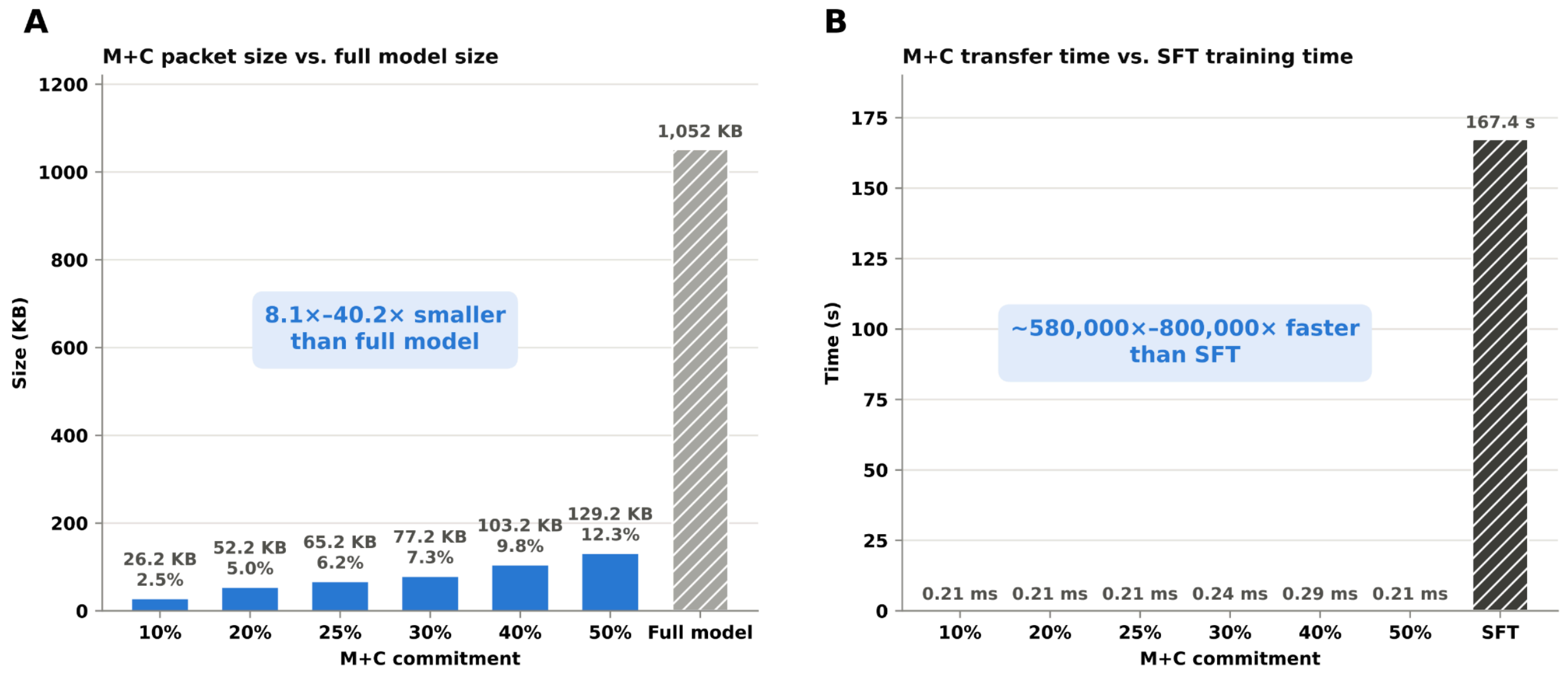

**Figure 8. M+C Compresses Model Data and Fast to Apply to Naive MLPs. A)** M+C packets at every commitment level are significantly reduced in size compared to the total size of the base MLP model. **B)** M+C overlays are rapid to apply, up to 0.29 milliseconds.

- **M+C Packets Compress Performance and Achieve Fast Application in MLP.** M+C packets range in size from 26.2 to 129.2 kilobytes (KB) (**Figure 8A**), representing a 40.2X to 8.1X reduction compared to the approximately 1,052 KB size of the MLP, respectively. In terms of timing to transfer M+C overlays, a range of 0.21 to 0.29 milliseconds, which is ~580K to ~800K faster than the ~167 seconds for supervised fine-tuning of the MLP (**Figure 8B**).

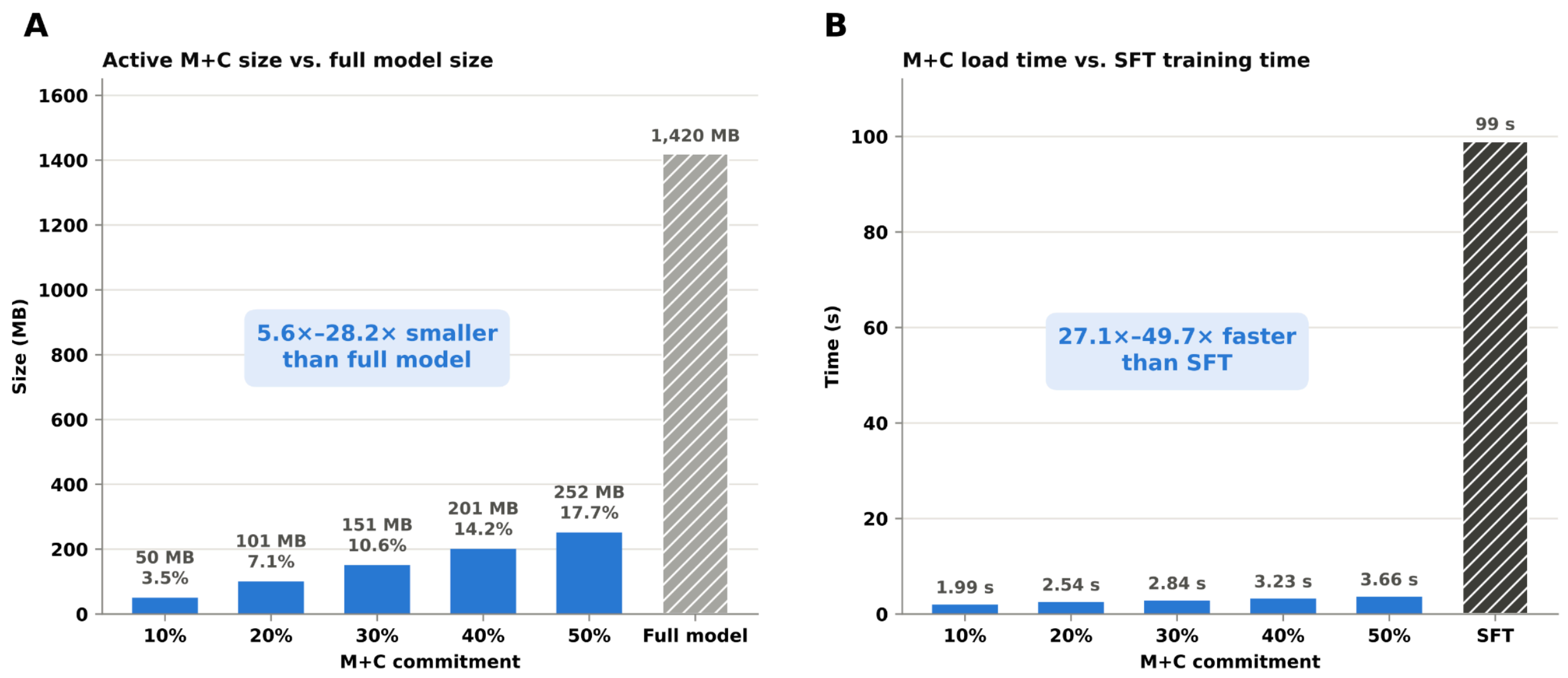

**Figure 9. M+C Compresses Model Data and Fast to Apply to Naive GPT-2 Instances. A)** M+C packets at every commitment level are significantly reduced in size compared to the total size of the base GPT-2 model. **B)** M+C overlays are rapid to apply, up to 3.66 seconds.

- **M+C Packets Compress Performance and Achieve Fast Application in GPT-2.** When we examined the M+C packets for GPT-2 medium, they range in size from 50 to 252 megabytes (MB) (**Figure 9A**), representing a 28.4X to 5.6X reduction compared to the approximately 1,420 MB size of the base model, resembling at the extreme, heavy pruning combined with quantization while still retaining substantial performance. For timing, application of M+C overlays ranged from 1.99 to 3.66 seconds, which is 27.1X to 49.7X faster than the 99 seconds for supervised fine-tuning of GPT-2 (**Figure 9B**).

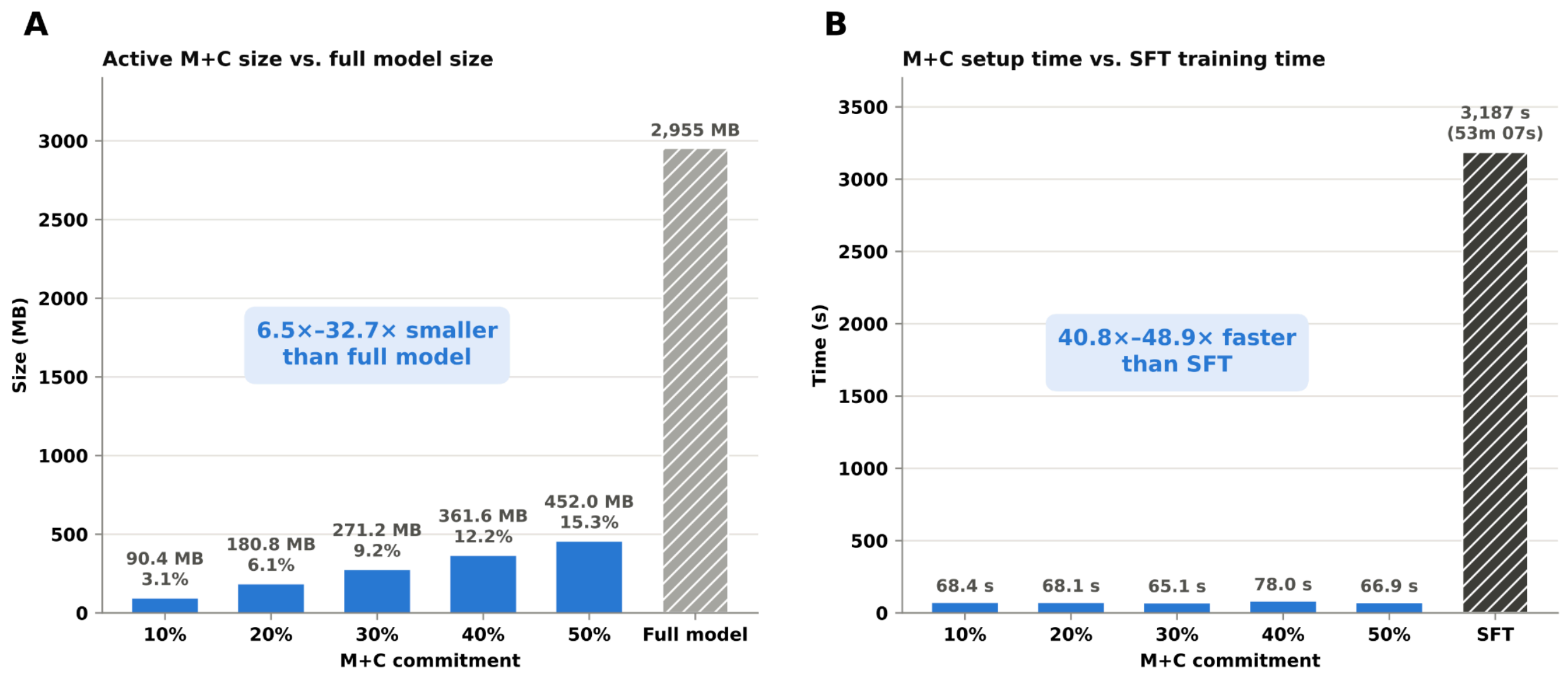

**Figure 10. M+C Compresses Model Data and Fast to Apply to Naive Qwen2.5 Instances. A)** M+C packets at every commitment level are significantly reduced in size compared to the total size of the base Qwen2.5 1.5B model. **B)** M+C overlays are rapid to apply, up to 78.0 seconds.

- **M+C Packets Compress Performance and Achieve Fast Application in Qwen2.5.** M+C packets range in size from 90.40 to 452 MB (**Figure 10A**), representing a 32.7X to 6.5X reduction, respectively, compared to the approximately 2,955 MB size of Qwen2.5 1.5B. This reduction resembles that of extreme model pruning or quantization, while still retaining substantial performance across all domains with virtually no performance loss (**Figure 10A**). In terms of timing to transfer M+C overlays, a range of 65.1 to 78.0 seconds, which is 40.8X to 48.9X faster than the 3,187 seconds for full supervised fine-tuning of Qwen2.5 (**Figure 10B**).

### 5.5 M+C Grants Training Privacy

***NURESA's M+C package confers resistance to reconstruction attacks and membership inference attacks for transformer models (GPT-2 and Qwen2.5) with M+C packets being indistinguishable from random chance for these attacks, even when the adversary is given access to the model's own embedding layer to project inputs into hidden space.***

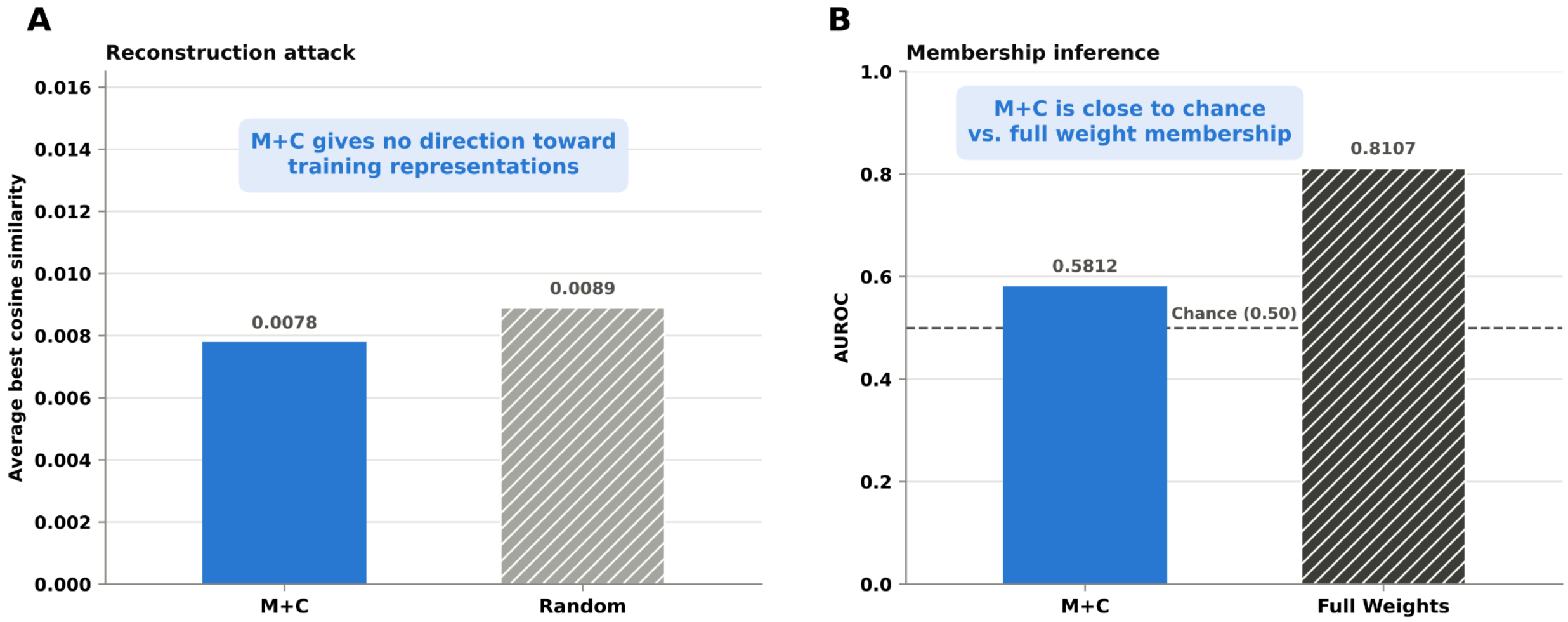

**Figure 11. M+C Confers Protection for Reconstruction Attack and Membership Inference Attack on GPT-2 Model. A)** Average cosine similarities to training representations are approximately the same for M+C packet vs. randomly selected vectors in the GPT-2 medium model. **B)** M+C has a near chance of indicating membership for training examples as compared to an adversary with full weight access to the GPT-2 model.

- **Privacy Capability of M+C Memory Transfer in GPT-2.** Given successful M+C transfer using GPT-2, we next sought to determine if an adversarial end user could get access to the data used to train the model with two different attack types. We held the commitment percentage at 30%, and after observing successful transfer (58.56 of M+C overlay vs. vanilla 245.79 on WikiText perplexity scores), inquired whether an M+C package could be used to reconstruct and recover training representations. We ran a gradient-based reconstruction attack, where we computed the cosine similarity between either M+C or random Gaussian vectors against a set of real training representations. We recorded the highest of these similarities, that is the closest match to any training example, reporting the average and standard deviation of these best cosine similarities. We saw that on average, M+C (0.0078) is relatively similar to random vectors (0.0089) **(Figure 11A)**, indicating an inability to recover training representations from the M+C packet. We then examined whether an adversary could determine if a data point was in the training set using only the M+C package. We trained a logistic regression classifier on standardized features from a random set of 70% of training examples, consisting of members and non-members, and evaluated it on the remaining 30%. Performance is reported as AUROC, or area under the receiver operating characteristic curve, indicating the probability that a randomly chosen member receives a higher membership score than a randomly chosen non-member. M+C had an AUROC of 0.5812, which is close to chance, vs. full weights access yielded AUROC of 0.8107 (**Figure 11B**), showing robust membership identification.

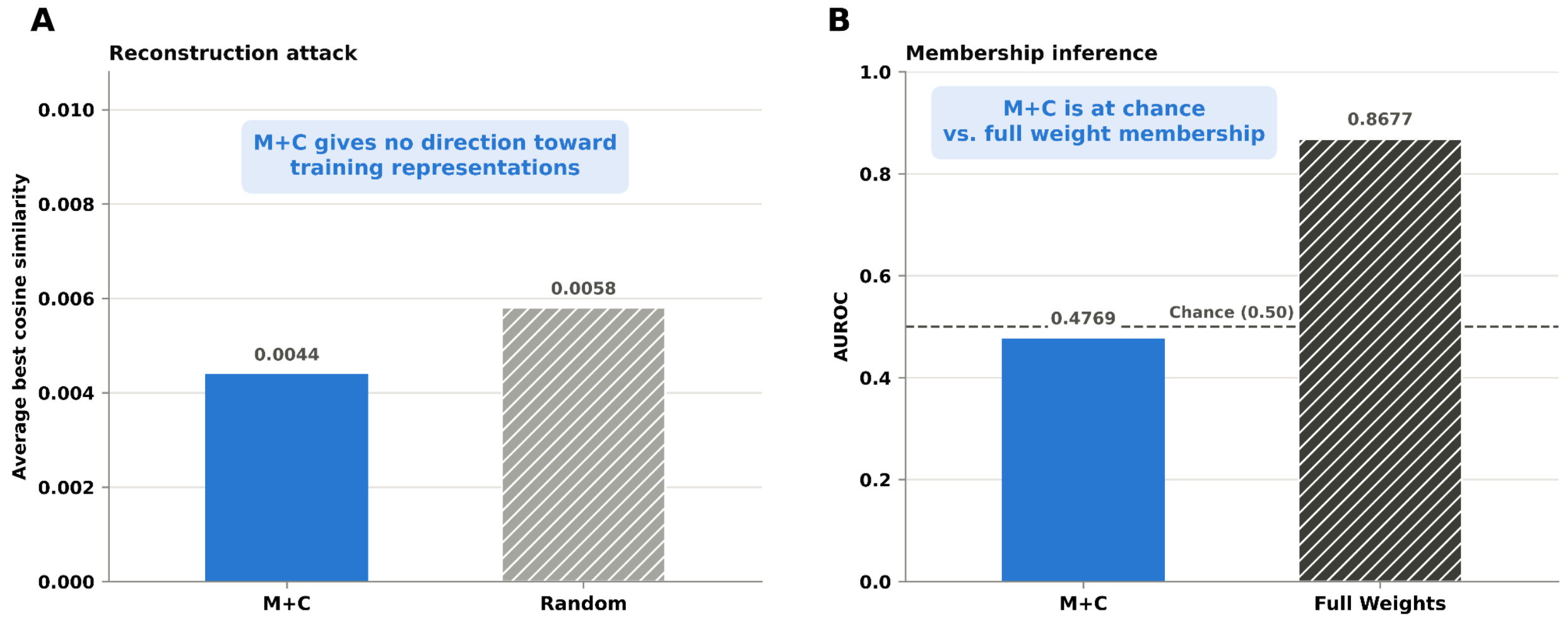

**Figure 12. M+C Confers Protection for Reconstruction Attack and Membership Inference Attack on Qwen2.5 Model. A)** Average cosine similarities to training representations are approximately the same for M+C packet vs. randomly selected vectors in the Qwen2.5 1.5B model. **B)** M+C has a near chance of indicating membership for training examples as compared to an adversary with full weight access to the Qwen2.5 model.

- **Privacy Capability of M+C Memory Transfer in Qwen2.5.** We next sought to determine if M+C privacy capability extends to Qwen2.5 1.5B model. We carried out similar reconstruction and membership attacks to those we conducted on GPT-2. We held the commitment percentage at 30%, and after observing successful transfer (56.0 of M+C overlay vs. trained at 61.0 on GSM8K accuracy scores), inquired whether an M+C package for Qwen2.5 could be used to reconstruct and recover training representations. We saw that M+C (0.0044) is similar to that of random noise (0.0058) on average of best cosine similarities **(Figure 12A**). We then examined whether an adversary could determine if a data point was in the training set using only the M+C package. M+C had an AUROC of 0.4769 vs. full weights access yielded AUROC of 0.8677 (**Figure 12B**), demonstrating M+C operates at chance on determining training set membership.

## 6 Conclusions and Future Work

We have presented NURESA, a neural layer that introduces persistent intra-unit memory to break the paradigm of parameter-locked capability. By decoupling computation from memory storage, NURESA achieves local adaptation without backpropagation and data-free capability transfer across MLPs and transformer models, i.e. GPT-2 and Qwen2.5. We showed that the NURESA M+C packet provided up to 1.5X to 4.5X performance to a fresh instance across all three models without further training at a file to full model ratio size reduction of values between 5.6X to 40.2X at a speed for transformer models ranging from 27.1X to 49.7X. We also saw that M+C provides training set privacy for GPT-2 and Qwen2.5 on two types of attacks, reconstruction and membership inference. ***Finally, of particular importance are two key findings: 1) M+C packets mitigated catastrophic forgetting in domains affected by SFT in both GPT-2 and Qwen2.5; and 2) M-selected neurons committed into M+C outperformed any other selection of neurons to recover performance gains.***

While NURESA introduces a novel paradigm for stateful neural execution, a few limitations frame ongoing research. The current implementation operates as a retrofitted register around a sample of key, standard architectures whose optimization routines assume parameter-locked capability. Also, in our experiments, we noticed that tuning the commitment threshold is critical; setting the threshold too low causes premature commitment into M+C, collapsing signal specificity. For the privacy experiments, we chose two attacks, one each for reconstruction and membership inference, but these do not necessarily completely rule out reconstruction or membership leakage of M+C. Finally, unit-level and agent-level transfer currently requires same-architecture, same-size source, and recipient models. Further research could cover multiple seeds of SFT, different types of SFT, as well as other model types. Another aspect to explore could be to combine M+C bundles to instill multiple skills to avoid supervised fine-tuning altogether or even surpass the SFT ceiling on a model's capability. Provocatively, could we develop different tools for creating and committing neurons, endowing them with memory of unique, shared features as recently excavated in simple artificial neurons by Cammarata and colleagues (Cammarata, 2026 [68])?

The observation that roughly ~30% of computational units capture task-critical capability aligns remarkably with biological findings where sparse engram cell populations (~10–30% of local neurons) store specific memory traces. In deep learning literature, the Lottery Ticket Hypothesis (Frankle & Carbin, 2019 [66]) demonstrated that sparse trainable subnetworks exist within dense models. However, lottery tickets identify non-transferable sparse weight masks that require retraining on raw data. NURESA M+C, by contrast, consolidates the underlying weight parameters into a portable register package that can be transplanted zero-shot across model instances. These observations also echo work on the Jacobian Space or J-Space (Anthropic, 2026 [67]), where vectors in this space align quite strongly with neuronal weights, and are a reduced subset of the model's footprint, mirroring a feature of M-selection for the NURESA M+C memory transfer register. In this way, M-selection and M+C broadly provide a built-in mechanism for interpretability to find and examine contents of high performing neurons.

We believe that NURESA is a "Day 1" step towards infinite and updateable working memory as an empirically valid, scalable method to copy memory states (Genewein et al., 2026 [69]). As mentioned in Genewein and colleagues work [69], memory state copying is a unique and underexploited property of AI systems, where not only their "DNA" (or as seen in neurobiological research program of Dr. Glanzman, genetic substrates like RNA (Bédécarrats et al., 2018 [16]), but also their "lifetime experience" as a memory state can be copied. This capability represents a key mechanism necessary to further recursive self-improvement through shared memory. We envision an evolutionary-like process by memory registers as reward functions to drive model improvement at scale with a native architecture built from the ground up around stateful memory units operating at biological efficiency, but with perfect memory state copying.

***We encourage the field to embrace methods to harness neuronal memory as a new and first primitive, and join us in taking steps toward infinite, persistent memory at scale.***

## Acknowledgments

We are very grateful to the colleagues who have provided valuable feedback with particular thanks to Dr. Seng-Beng Ho for looking at the manuscript draft and providing suggestions.

## Competing Interests

Alain C. Glanzman and David L. Glanzman are co-inventors on a patent application (UCLA Case No. 2025-328) covering aspects of the NURESA architecture, co-assigned to The Regents of the University of California and Alain C. Glanzman.

## Data and Code Availability

Data and code availability requests related to NURESA should be directed to UCLA Technology Development Group (joel.kehle@tdg.ucla.edu). Due to NURESA's patent status, UCLA TDG is required to adjudicate access.

## AI Use in Manuscript Preparation

During the preparation of this manuscript, the authors used language models to polish wording and draft sentences as well as critique the paper's core argument, reference searches, and reference management. The authors reviewed and edited all output and take full responsibility for the content of this publication.